\documentclass{article}

\usepackage{microtype}
\usepackage{graphicx}
\usepackage{subfigure}
\usepackage{booktabs} 

\usepackage{amssymb,amsmath,xcolor}
\usepackage{bm}
\usepackage{csquotes}
\usepackage{natbib}

\usepackage{wrapfig}

\usepackage[draft,inline,nomargin,index]{fixme}

\fxsetup{theme=color,mode=multiuser,inlineface=\itshape,envface=\itshape}
\FXRegisterAuthor{sv}{asv}{\colorbox{gray!10!white}{\color{black}Suresh}}
\FXRegisterAuthor{sf}{asf}{\colorbox{blue!10!white}{\color{black}Sorelle}}
\FXRegisterAuthor{cs}{acs}{\colorbox{red!10!white}{\color{black}Carlos}}
\FXRegisterAuthor{lk}{alk}{\colorbox{purple!10!white}{\color{black}Lizzie}}
\fxusetargetlayout{color}

\usepackage{hyperref}

\usepackage[accepted]{icml2020}

\icmltitlerunning{Problems with Shapley-value-based explanations as feature importance measures}

\begin{document}

\twocolumn[
\icmltitle{Problems with Shapley-value-based explanations as feature importance measures}



\icmlsetsymbol{equal}{*}

\begin{icmlauthorlist}
\icmlauthor{I. Elizabeth Kumar}{u}
\icmlauthor{Suresh Venkatasubramanian}{u}
\icmlauthor{Carlos Scheidegger}{az}
\icmlauthor{Sorelle A. Friedler}{ha}
\end{icmlauthorlist}

\icmlaffiliation{u}{School of Computing, University of Utah, Salt Lake City, UT, USA}
\icmlaffiliation{az}{Department of Computer Science, University of Arizona, Tucson, AZ, USA}
\icmlaffiliation{ha}{Department of Computer Science, Haverford College, Haverford, PA, USA}

\icmlcorrespondingauthor{I. Elizabeth Kumar}{kumari@cs.utah.edu}

\icmlkeywords{SHAP, Shapley values, explainable machine learning}

\vskip 0.3in
]



\printAffiliationsAndNotice{} 

\begin{abstract}
Game-theoretic formulations of feature importance have become popular as a way to ``explain" machine learning models. These methods define a cooperative game between the features of a model and distribute influence among these input elements using some form of the game's unique Shapley values. Justification for these methods rests on two pillars: their desirable mathematical properties, and their applicability to specific motivations for explanations. We show that mathematical problems arise when Shapley values are used for feature importance, and that the solutions to mitigate these necessarily induce further complexity, such as the need for causal reasoning. We also draw on additional literature to argue that Shapley values are not a natural solution to the human-centric goals of explainability.

\end{abstract}

\section{Introduction}

Machine learning models are increasingly being used to replace human decision-making for tasks involving some kind of prediction. As state-of-the-art predictive machine learning models become increasingly inscrutable, there has been an increase in concern that the black-box nature of these systems can obscure undesirable properties of the decision algorithm, such as illegal bias or signals accidentally learned from artifacts irrelevant to the task at hand. More recently, attempts have been made to ``explain" the output of a complicated function in terms of its inputs to address these and other concerns. One of the more prominent tools in this literature has been the Shapley value, a method for additively attributing value among players of a cooperative game. In this setting, the ``players" are the features used by the model, and the game is the prediction of the model. A variety of methods to assign feature influence using the Shapley value have recently been developed \citep{lipovetsky2001analysis, vstrumbelj2014explaining, lundberg2018consistent, datta2016indirect, merrick2019explanation, frye2019asymmetric, aas_explaining_2019}.

In this paper, we demonstrate that Shapley-value-based explanations for feature importance fail to serve their desired purpose in general. We make this argument in two parts. Firstly, we show that applying the Shapley value to the problem of feature importance introduces mathematically formalizable properties which may not align with what we would expect from an explanation. Secondly, taking a human-centric perspective, we evaluate Shapley-value-based explanations through established frameworks of what people expect from explanations, and find them wanting. We find that the game theoretic problem formulation of Shapley-value-based explanations do not match the proposed use cases for its solution, and thus caution against their usage except in narrowly constrained settings where they admit a clear interpretation.

We describe the different Shapley-value-based explanation frameworks in Section~\ref{sec:background}, and present our two-part critique in Sections~\ref{sec:mathematical-issues} and \ref{sec:applicability-issues}. We discuss these results and provide some suggestions in Section~\ref{sec:concl}. 

\section{Background}
\label{sec:background}
In this section, we define the Shapley value and articulate the different ways in which it has been applied to the problem of feature importance. 

\subsection{Classical Shapley values}
\label{subsec:classic}
In cooperative game theory, a coalitional game consists of a set of $N$ players and a characteristic function $v$ which maps subsets $S \subseteq \{1, 2, ..., N\}$ to a real value $v(S)$, satisfying $v(\emptyset) = 0$. The value function represents how much collective payoff a set of players can gain by ``cooperating" as a set. The Shapley value is one way to allocate the total value of the grand coalition, $v(\{1, 2, ..., N\})$, between the individual players. It is based on trying to answer the question: how much does player $i$ contribute to the coalition?

The marginal contribution $\Delta_v(i, S)$ of player $i$ with respect to a coalition $S$ is defined as the additional value generated by including $i$ in the coalition:
\begin{equation}
\Delta_v(i, S) = v(S \cup i) - v(S)
\end{equation}
Intuitively, the Shapley value can be understood as a weighted average of a player's marginal contributions to every possible subset of players. Let $\Pi$ be the set of permutations of the integers up to $N$, and given $\pi \in \Pi$ let $S_{i,\pi} = \{j: \pi(j) < \pi(i)\}$ represent the players preceding player $i$ in $\pi$. The \emph{Shapley value} of player $i$ is then \begin{equation}
\phi_v(i) = \frac{1}{N!} \sum_{\pi \in \Pi} \Delta_v(i, S_{i,\pi})
\end{equation}
This can be rewritten in terms of the unique subsets $S \subseteq \{1, 2, ..., N\}$ and the number of permutations for which some ordering of $S$ immediately precedes player $i$: \begin{equation}
\phi_v(i) = \frac{1}{N!} \sum_{S \subseteq \{1, 2, ..., N\}} |S|! (N-|S|-1)! \Delta_v(i, S)
\end{equation}
This value is the unique allocation of the \emph{grand coalition} $v(\{1, 2, ..., N\})$ which satisfies the following axioms:

\textbf{Symmetry}: For two players $i,j$, if $\Delta_v(i, S) = \Delta_v(j, S)$ for any subset of players $S$, then $\phi_v(i) = \phi_v(j)$.

\textbf{Dummy}: For a single player $i$, if $\Delta_v(i, S) = 0$ for all subsets $S$, then $\phi(i) = 0$.

\textbf{Additivity}: For a single player $i$ and two value functions $v$ and $w$, $\phi_v(i) + \phi_w(i) = \phi_{v+w}(i)$.

\subsection{Shapley values for feature importance}

Several methods have been proposed to apply the Shapley value to the problem of feature importance. Given a model $f(x_1, x_2, ..., x_d)$, the features from 1 to $d$ can be considered players in a game in which the payoff $v$ is some measure of the importance or influence of that subset. The Shapley value $\phi_v(i)$ can then be viewed as the ``influence" of $i$ on the outcome.

In this section, we describe methods which consist of defining a value function $v_f$ with respect to a model $f$, and computing (or approximating) the resulting Shapley values. We will use the following notation:

$D$: the set of features $\{1, 2, ..., d\}$

$\bm{X}$: a multivariate random variable $\{X_1, X_2, ..., X_d\}$

$\bm{x}$: a set of values $\{x_1, x_2, ..., x_d\}$

$\bm{X}_S$: the set of random variables $\{X_i: i \in S\}$

$\bm{x}_S$: the set of values $\{x_i: i \in S\}$

\subsubsection{Value functions}
\label{subsubsec:value}
Shapley values have a fairly long history in the context of feature importance. \citet{kruskal1987relative} and \citet{lipovetsky2001analysis} proposed using the Shapley value to analyze global feature importance in linear regression by using the value function $v_f(S)$ to represent the $R^2$ of a linear model $f$ built on predictors $S$, to decompose the variance explained additively between the features. \citet{owen2017shapley} applied the Shapley value to the problem of sensitivity analysis, where the total variance of a function is the quantity of interest.

Many recently proposed ``local" methods \cite{ribeiro2016should,lundberg2017unified,lundberg2018consistent}
define a value function $v_{f,x}: 2^d \rightarrow \mathbb{R}$ that depends on a specific data instance $\bm{x}$ to explain how each feature contributes to the output of the function on this instance. The value of the grand coalition, in this setting, is the prediction of the model at $\bm{x}$: $v_{f,x}(D) = f(\bm{x})$.
In addition, to use Shapley values as an ``explanation" of the (grand coalition of) features in this way, these methods also need to specify how $v_{f,x}$ acts on proper subsets of the features.

%


The definitions of \emph{Shapley sampling values} \cite{vstrumbelj2014explaining}, as well as \emph{SHAP values} \cite{lundberg2017unified}, are derived from defining $v_{f,x}(S)$ as the \emph{conditional} expected model output on a data point when only the features in $S$ are known:
\begin{equation}
v_{f,x}(S) = E[f(\bm{X})|\bm{X}_S = \bm{x}_S] = E_{X_{\bar{S}}|X_S} [f(\bm{x}_S, \bm{X}_{\bar{S}})]
\end{equation}

Quantitative Input Influence (QII) \cite{datta2016algorithmic} draws on ideas from causal inference to propose simulating an \emph{intervention} on the features not in $S$, thus breaking correlations with the features in $S$:
\begin{equation}
v_{f,x}(S) = E_{\mathcal{D}}[f(\bm{x}_S, \bm{X}_{\bar{S}})]
\end{equation}
where the distribution $\mathcal{D}$ is derived from the product of the marginal distributions of the features in $\bar{S}$. The approach of using a distribution other than that of the original data was further generalized by \cite{merrick2019explanation}, who also propose the Formulate, Approximate, Explain (FAE) framework, so as to unify a number of different approaches to Shapley value explanations.

\begin{table*}[htbp]
\centering
\begin{tabular}{ |c|c|c| } 
 \hline
 Method & $v_{f,x}(S)$ & $\hat{v}_{f,x}(S)$\\ 
 \hline
 KernelSHAP, Shapley sampling values & $E_{X_{\bar{S}}|X_S} [f(\bm{x}_S, \bm{X}_{\bar{S}})]$ & $E_{\mathcal{D}} [f(\bm{x}_S, \bm{X}_{\bar{S}})]$\\
 QII, FAE, Interventional TreeSHAP & $E_{\mathcal{D}}[f(\bm{x}_S, \bm{X}_{\bar{S}})]$ & $E_{\mathcal{D}}[f(\bm{x}_S, \bm{X}_{\bar{S}})]$ \\ 
 Conditional TreeSHAP, \citet{frye2019asymmetric}, \citet{aas_explaining_2019} & $E_{X_{\bar{S}}|X_S} [f(\bm{x}_S, \bm{X}_{\bar{S}})]$ &  $E_{X_{\bar{S}}|X_S} [f(\bm{x}_S, \bm{X}_{\bar{S}})]$ \\

 \hline
 \end{tabular}
\caption{\label{tab:table1} Proposed value function $v_{f,x}$ for each method, compared with the quantity $\hat{v}_{f,x}$ the algorithm actually approximates. The interventional distribution ${\mathcal{D}}$ used depends on the method (i.e., for KernelSHAP it is the observational joint distribution of $\bar{X}$). }
\end{table*}

\subsubsection{Algorithms}

Methods based on the same value function can differ in their mathematical properties based on the assumptions and computational methods employed for approximation. 
TreeSHAP \cite{lundberg2018consistent}, an efficient algorithm for calculating SHAP values on additive tree-based models such as random forests and gradient boosting machines, can estimate $E_{X_{\bar{S}}|X_S} [f(\bm{x}_S, \bm{X}_{\bar{S}})]$ by observing what proportion of the samples in the training set matching the condition $\bm{x}_S$ fall into each leaf node, a method which does not rely on a feature independence assumption.
In the algorithm for KernelSHAP \cite{lundberg2017unified}, conditional expectations are estimated by assuming feature independence; samples of the features in $\bar{S} = D \setminus S$ are drawn from the \emph{marginal} joint distribution of these variables. This effectively approximates an expectation over an \emph{interventional} distribution instead, though in a slightly different way from QII.


In Table \ref{tab:table1}, we categorize each method based on how they \emph{define} a value function $v_{f,x}(S)$ and how they \emph{estimate} that value function $\hat{v}_{f,x}(S)$. In the rest of the paper, we will refer to these value functions as either interventional or conditional based on the estimation method. That is to say, KernelSHAP, Shapley sampling values, QII, and FAE are \emph{interventional} methods, while TreeSHAP as well as some other algorithms we will introduce later are \emph{conditional}.

\section{Mathematical issues}
\label{sec:mathematical-issues}

We now present a number of mathematically articulated problems that arise when we attempt to interpret Shapley values as feature importance measures. These problems arise from the estimation procedures that are in use as well as the fundamental axiomatic structure of Shapley values. 

\subsection{Conditional versus interventional distributions}

A fundamental difference between the interventional and conditional value functions is revealed by what we call the \emph{indirect influence} debate. Suppose $f$ is defined with domain $\mathbb{R}^d$, but for a certain feature $i$, $f(\bm{x}) = f(\bm{x}')$ whenever $x_j = x_j'$ for all $j \neq i$; that is to say, intervening on the value of $x_i$ alone does not change the output of $f$. We call this a variable with \emph{no interventional effect.}

Should a feature with no interventional effect be considered an ``input" to this function? We could define a new function $f'$ with domain $\mathbb{R}^{d-1}$ to perfectly capture the output, so perhaps not. What if, in the relevant input space, $x_i$ is a statistical proxy for some $x_j$ which \emph{does} affect the output of $f$? Shapley value based feature importance methods must grapple with these choices.

\citet{adler2018auditing} take the information-theoretic position that ``the information content of a feature can be estimated by trying to predict it from the remaining features." This perspective can help diagnose situations where an undesirable proxy variable is being used by a model, as in the classic case of redlining.
While \citeauthor{adler2018auditing} go on to analyze how the \emph{accuracy} of a model depends on indirect information, the \emph{conditional} value function aligns with this information-theoretic principle as well: If a certain feature $i$ can help predict the features in $\bar{S}$, then the quantities $v_{f,x}(S \cup i) = E [f(\bm{X})|\bm{X}_{S \cup i} = \bm{x}_{S \cup i}]$ and $v_{f,x}(S) = E [f(\bm{X})|\bm{X}_S = \bm{x}_{S}]$ may be meaningfully different, meaning that the marginal contribution of feature $i$ is nonzero. For this reason the Shapley value of the conditional value function may attribute influence to features with no interventional effect, a positive thing from the perspective of \citeauthor{adler2018auditing}.

\citet{merrick2019explanation}, on the other hand, criticize the capacity to attribute indirect influence as being paradoxical, and show that interventional methods will \emph{never} attribute attribute influence to an $x_i$ which has no interventional effect on $f$, which they see as a desirable property.

Unfortunately, the decision between the two types of value functions is a catch-22. Both methods introduce serious issues: Choosing a conditional method requires further modeling of how the features are interrelated, which we describe in \ref{sec:issu-with-cond}, while choosing an interventional method induces an ``out-of-distribution" problem which we address in \ref{sec:issu-with-interv}.



\subsubsection{Issues with conditional distributions}
\label{sec:issu-with-cond}
The conditional value function induces two major difficulties. First, the exact computation of the Shapley value for a conditional value function would require knowledge of $2^d$ different multivariate distributions, and so a significant amount of approximation or modeling is necessary. Second, since influence can be computed on an arbitrarily large set of features, it becomes necessary to choose a set that is meaningful because the explanations may change based on which features are considered.

Solutions have been proposed to deal with the computational complexity of this problem. The TreeSHAP algorithm estimates the conditional expectations of any tree ensemble directly, without sampling, using information computed during model training. The algorithm utilizes information about the training instances which fall into each leaf node to model each conditional distribution. It is \emph{not}, however, set up to attribute influence to variables without an interventional effect, as the trees contain no information about the distribution of variables not in the model.

For arbitrary types of models, estimating the conditional expectations requires a substantial amount of additional modeling of relationships in the data which are not necessarily captured by the model that one is trying to explain. \citet{aas_explaining_2019} and \citet{frye2019asymmetric} have developed methods that aim to generate in-distribution samples for the relevant calculations. 

Even if computational issues are resolved, there are additional inconsistencies introduced by the capacity of the Shapley value to attribute influence to an arbitrarily large feature set given a single function. The modeler must decide which features count as players in the cooperative game and which are redundant, and since the problem definition posits that the attributions add up to the value of $f(x)$, this choice can affect the resulting explanations.

Consider the addition of a redundant variable $C$ to a dataset with two features, $A$ and $B$, so that $P(X_C = X_B) = 1$. Suppose a model $f$ is trained on all three features. Intuitively, the features $B$ and $C$ should be equally informative to the model and so should have the same Shapley value under the conditional value function. Formally, the following properties will hold:

\begin{align}E [f(\bm{X})|X_B, X_C] &= E [f(\bm{X})|X_B] \\ &= E [f(\bm{X})|X_C] \\
E [f(\bm{X})|X_A, X_B, X_C] &= E [f(\bm{X})|X_A, X_B] \\ &= E [f(\bm{X})|X_A, X_C]\end{align}

so this means $v_{f,x}(B) = v_{f,x}(C) = v_{f,x}(BC)$ and $v_{f,x}(AB) = v_{f,x}(AC) = v_{f,x}(ABC)$. Therefore, for any data instance $x$,
\begin{align}
\phi_v(A)  
&= \frac{1}{3}\Delta_v(A, \emptyset) + \frac{2}{3}\Delta_v(A, BC) \\
\phi_v(B) = \phi_v(C) 
&= \frac{1}{3}\Delta_v(B, \emptyset) + \frac{1}{6}\Delta_v(B, A)
\end{align}

Now consider what would happen if we defined a new function $f'(x_A, x_B) = f(x_A, x_B, x_B)$. For any data instance, since $x_B = x_C$, $f'(\bm{x}) = f(\bm{x})$. It is effectively the same model for all in-distribution data points, so the games $v_{f,x}$ and $v_{f', x}$ are the same for all subsets of variables. Yet if we choose to limit the scope of our explanation to two variables instead of three, the attribution for both $A$ and $B$ will come out to be different:

\begin{align}
\phi'_v(A)  
&= \frac{1}{2}\Delta_v(A, \emptyset) + \frac{1}{2}\Delta_v(A, BC)
\end{align}
\begin{align}
\phi'_v(B)  
&= \frac{1}{2}\Delta_v(B, \emptyset) + \frac{1}{2}\Delta_v(B, A)
\end{align}

Notice that $\phi'_v(B)$ is neither equal to $\phi_v(B)$, its assigned influence in the 3-variable setting, nor $\phi_v(B) + \phi_v(C)$, the ``total" influence of the two identical variables in the 3-variable setting. The relative apparent importances of $A$ and $B$ thus depend on whether $C$ is considered to be a third feature, even though the two functions are effectively the same. 

It is not obvious whether two statistically related features should be considered as separate ``players" in the cooperative game, yet this choice has an impact on the output of these additive explanation models. Suppose, for instance, that $B$ is a sensitive feature, and $C$ is a non-sensitive feature that happens to perfectly correlate with it. Two different ``fairness" audits of the same function would come out with quantitatively different results.

\citet{frye2019asymmetric} propose to a solution to the problem in terms of incorporating causal knowledge:

\begin{quote}
...If $x_i$ is known to be the deterministic causal ancestor of $x_j$ , one might want to attribute all the importance to $x_i$ and none to $x_j$.
\end{quote}

They propose not only discounting fully redundant variables which are causal descendants of other variables in the model, but relaxing the symmetry axiom which uniquely defines the Shapley value. Instead of averaging marginal contributions over every permutation, they suggest defining a quasivalue which considers only certain permutations; for example, orderings which place causal ancestors before their descendants.

In this framework, fully redundant features will receive zero attribution \emph{and} will not change the resulting value of the remaining features. For instance, in the above example, if variable $C$ were known to be a causal descendant of $B$, the Asymmetric Shapley Values of $A$ and $B$ under $f'$ will be the same as they were under $f$.

A fully specified causal model is not required to use this method: they ``span the data-agnosticism continuum in the sense that they allow any knowledge about the data, however incomplete, to be incorporated into an explanation of the model’s behaviour." The results in \citet{frye2019asymmetric} demonstrate, however, the \emph{sensitivity} of the game theoretic approach to the amount of prior knowledge about the relative agency of each feature, which we consider a significant limitation of the approach.

\begin{figure*}
  \centering
  \includegraphics[height=2.5in]{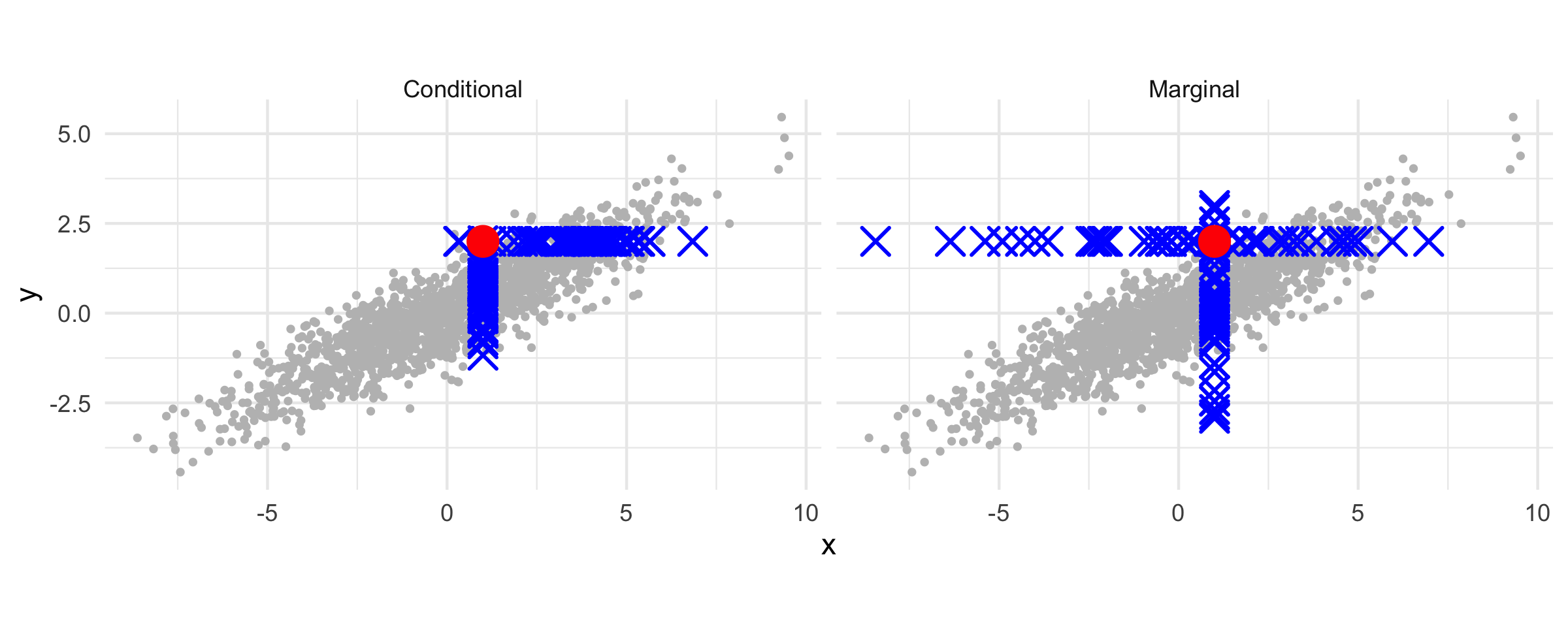}
    \vspace{-2.2em}
  \caption{Samples that might be drawn to estimate $E[f(1,Y)]$ and $E[f(X,2)]$ to explain $f(1,2)$ for some function $f$, given correlated Gaussian distributions for $X$ and $Y$, depending on whether the expectation is taken over $X|Y=2$ and $Y|X=1$ (left) or $X$ and $Y$ (right)}
  \label{fig:ood}
  
\end{figure*}

There are thus both practical and epistemological challenges with computing the Shapley values of games with a conditional value function. 

\subsubsection{Issues with interventional distributions}
\label{sec:issu-with-interv}

Conditional value functions introduce undesirable complexities to the feature importance problem, so those inclined against methods with the capacity for attributing indirect influence may prefer the methods interventional value functions instead. These methods, however, are highly sensitive to properties of the model which are not relevant to what it has learned about the data it was trained on.

Methods which use an interventional value function fundamentally rely on evaluating a model on \emph{out-of-distribution} samples (Figure \ref{fig:ood}). Consider, for example, a model trained on a data set with three features: $X_1$ and $X_2$, both $N(0,1)$, and an engineered feature $X_3 = X_1 X_2$. To calculate $v_{f,x}(\{1,2\})$ for some $x = \{x_1, x_2, x_3\}$, we would have to estimate $E(f(x_1, x_2, X_3))$ over some distribution for $X_3$ which does not depend on $x_1$ or $x_2$. Therefore we will almost certainly have to evaluate $f$ on some sample $\{x_1, x_2, x'_3\}$ which \emph{does not respect} $x'_3 = x_1 x_2$ - thus, it is well \emph{outside} the domain of the actual data distribution. The model $f$ has never seen an example like this in training, and has therefore not learned much about this part of the feature space. Its predictions on this feature space are not necessarily relevant to the task of explaining an in-distribution sample, yet the explanations will be affected by them.

This ``out-of-distribution" phenomenon has been explored recently by \citet{pleasestop}, who show why ``permutation-based" methods to evaluate feature importance can be highly misleading: when values are substituted into feature set $\bar{S}$ that are unlikely or impossible when conditioned on feature set $S$, the model $f$ is forced to \emph{extrapolate} to an unseen part of the feature space. They show that these feature importance methods are highly sensitive to the way in which the model extrapolates to these edge cases, which is undesirable information for a model ``explanation" to capture.

\citet{SlackHilgard2020FoolingLIMESHAP} demonstrate how to exploit this sensitivity by devising models which illegally discriminate on some protected feature for in-distribution samples, but exhibit different behavior on the out-of-distribution samples used by KernelSHAP so as to simulate ``fairness" in the resulting explanations. By manipulating the model's behavior on unfamiliar parts of the feature space, they can twist the explanations on the familiar part to their will.


%

These challenges illustrate that intervening on a subset of features of a data case before applying a model trained on a sample from a certain distribution is inherently misleading. 

\subsection{Additivity constraints}
\label{sec:addit-constr}

In addition to the problems demonstrated above, which have to do with the choice between two families of value functions, we also identify problems which are common to both. These are linked to the  axiomatic underpinnings of Shapley values.

For any two of the axioms described in Section~\ref{subsec:classic}, there exists an alternative attribution between players which satisfies those two but not the other; the Shapley value is therefore only unique because it satisfies all three. Since the notion of the sum of two games is not especially meaningful, the \textbf{Additivity} axiom has been described by game theorists as ``mathematically convenient" and ``not nearly so innocent as the other two" \cite{osborne1994course}. The choice to constrain the value to be unique in this way has implications for what kinds of models can be explained intuitively by the Shapley value. Even in simple cases where feature independence renders the interventional versus conditional debate irrelevant, we find the Shapley value conceptually limited for non-additive models.

The Shapley value seems to intuitively align with what is considered important in an additive setting. Consider applying any of the expectation value functions to $f(x) = \beta_0 + \beta_1 x_1 + ...  + \beta_d x_d$ where the features $X_i$ are independent. For any subset $S$,
\begin{align}
    v_{f,x}(S) &= E_{X_{\bar{S}}|X_S} [f(\bm{x}_S, \bm{X}_{\bar{S}})] \\
    &= f(\bm{x}_S, E[\bm{X}_{\bar{S}}])  \\
    &= \sum_{j \in S} \beta_j x_j + \sum_{j \in \bar{S}} \beta_j E[X_j]+ \beta_0
\end{align}

so the marginal contribution for feature $i \not \in S$ is
\begin{align}
    &\sum_{j \in S \cup i} \beta_j x_j + \sum_{j \in \bar{S \cup i}} \beta_j E[X_j] + \beta_0 \notag\\
    &- \left(\sum_{j \in S} \beta_j x_j + \sum_{j \in \bar{S}} \beta_j E[X_j] + \beta_0\right) \\
    =&\sum_{j \in S} \beta_j x_j + \beta_i x_i +  \sum_{j \in \bar{S \cup i}} \beta_j E[X_j]+ \beta_0 \notag\\
    &- \left(\sum_{j \in S} \beta_j x_j + \beta_i E[X_i] + \sum_{j \in \bar{S \cup i}} \beta_j E[X_j] + \beta_0\right) \\
    =& \beta_i(x_i - E[X_i])
\end{align}



In this way, the Shapley value is supported by the common intuition that coefficient size, if variables are appropriately scaled, signals importance in a linear model.

The additivity axiom is aligned with additive models in another way: the games resulting from two models sum to the expectation game of the sum of the two models. This seems reasonable when the models are additive in the first place.

Now imagine if the additivity constraint were relaxed. We could use an alternative attribution $\psi$ which satisfies the other two axioms: $\psi: \psi(i) = v(i)$ for $i \in U$ and $\psi(i) = \frac{1}{|U|}(v(D) - \sum_{j \in U} v(j))$  where $U$ is the set of dummy features. Using the expectation value function in this setting, any feature which did not satisfy $\beta_i(x_i - E[X_i]) = 0$ would get \emph{the same attribution.} In this sense the additivity constraint seems necessary for a game-based feature attribution to provide any meaningful quantities about an additive model. Under an interventional interpretation of the attribution --- using the values to assess which data changes produce the largest model prediction change --- this is not a helpful property.

Under an interventional interpretation, Shapley values are as uninformative for  non-additive models as this alternative attribution is for linear ones. For instance, any value function which always evaluates to 0 except on the grand coalition will evenly distribute influence among players. Consider a model given by $f(x) = \Pi_{j=1}^d x_d$ where the features are independent and centered at 0. Then for any subset $S$,
\begin{align}
    v_{f,x}(S) &= E [f(\bm{X}_S, \bm{X}_{\bar{S}})|\bm{X}_S = \bm{x}_S] \\
    &= E[\prod_{j=1}^d X_d | \bm{X}_S = \bm{x}_S] \\
    &= \prod_{j=1}^d E[X_j | \bm{X}_S = \bm{x}_S] \\
    &= \left(\prod_{j \in S} x_j\right) \left(  \prod_{j \in \bar{S}} E[x_j]\right)
\end{align}

which, since $E[x_j]$ is 0, is always 0 unless $S = D$. Then the Shapley value for every feature $i$ is $\frac{1}{d} f(x)$, regardless of the value $x_i$. Even if, for instance, the magnitude of one of the variables is much higher than the other. This property will, in fact, hold for \emph{all multiplicative functions} of independently distributed, zero-centered data.

Shapley values are touted for their ``model-agnostic" quality, but under the lens of a particular interpretation, this is not the case.

\section{Human-centric issues}
\label{sec:applicability-issues}

The analysis from Section~\ref{sec:mathematical-issues} demonstrates the mathematical issues with feature importance methods derived from Shapley values and suggests how one might mitigate them.
In this section we turn to the human side of the interaction between feature importance methods and the people who use them. This perspective is closer in spirit to the ``human-grounded metrics'' that \citet{doshi-velez_towards_2017} describe in comparison with the ``functionally-grounded evaluation'' of the previous section.


We use the framework set out by \citet{selbst_intuitive_2018}, who argue that there are three general motivations behind the call for explanations in AI.

\begin{quote}
The first is a fundamental question of autonomy, dignity, and personhood.  The second is a more instrumental value:  educating the subjects of automated decisions about how to achieve different results.  The third is a more normative question—the idea that explaining the model will allow people to debate whether the model’s rules are justifiable.
\end{quote}

In this section, we attempt to reconcile the Shapley value feature importance formalization of machine learning ``explanations" with these three goals. We argue that the theoretical properties of the Shapley value are not naturally well-suited to any one of these objectives. While we focus here on these issues in the context of Shapley values, many of these critiques also apply to other explanatory methods.

\subsection{Explanations as contrastive statements}

The presence of the phrase ``right to explanation" in the GDPR illustrates the sense many of us have that it is inherently unethical to make decisions about an individual without providing an explanation, in a way that \citet{selbst_intuitive_2018} argue has more to do with ``procedural justice" than ``wanting an explanation for the purpose of vindicating certain specific empowerment or accountability goals.''

It is not immediately clear how to formally evaluate a method that provides explanations merely because it should, rather than to improve on a particular metric or task. In this setting, \citeauthor{doshi-velez_towards_2017} suggest the empirical approach of running user tests where humans are provided with explanations and they evaluate their ``quality". But in fact, what humans consider a good explanation has been studied extensively in the social sciences, leading to several formal theories of how humans generate and select explanations.

\citet{miller2019explanation} provides an overview of this literature. One of his major findings is that the way humans explain phenomena to each other is through  \emph{contrastive} statements: 

\begin{displayquote}
People do not explain the causes for an event per se, but explain the cause of an event relative to some other event that did not occur; that is, an explanation is always of the form ``Why P rather than Q?", in which P is the target event and Q is a counterfactual contrast case that did not occur.
\end{displayquote}

He attributes this insight to work by \citet{lipton1990contrastive}. More recently, a similar argument has been made by \citet{merrick2019explanation}, referencing earlier work by \citet{kahneman1986norm}. 

We now outline different ways in which Shapley values can be interpreted as contrastive explanations.

\subsubsection{Shapley value sets as a single contrastive statement}
\label{sec:shapley-value-sets}

The above-mentioned research supports the hypothesis that people ask for explanations when the outcome, P, is ``unexpected" compared to the outcome Q. In this sense, we can interpret Shapley-based explanations as a contrastive statement where the outcome to be explained is $v(D)$ and the foil -- the counterfactual case which did not happen -- is implicitly set to be $v(\emptyset)$. 
In the ``local" settings described earlier, $v(D)$ is $f(x)$ and $v(\emptyset)$ is $E(f(x))$:
\[f(x) = E(f(x)) + \phi_1 + \phi_2 + ... + \phi_d\]
Thus, the Shapley values can be thought of as a set of answers to the question, ``Why $f(x)$ rather than $E(f(x))$?"

While the expected value of a function seems like a natural foil to an ``unexpected" $f(x)$, due to the properties of the expectation, there may not be a scenario in the data space of $X$ with the outcome $E(f(x))$. Thus, the expected value may not be ``expected" by anyone with a reasonable understanding of the situation at hand at all.

If we are willing to consider intervention distributions (Section \ref{subsubsec:value}), then the framework provided by \citet{merrick2019explanation} provides a slightly different contrastive explanation: in their setting, the Shapley value assignment can be thought of as a set of answers to the question, ``Why $f(x)$ rather than $f(r)$?'', where $r$ is chosen from the reference distribution. This of course requires the specification of the reference distribution and carries with it the estimation issues described above in Section~\ref{sec:issu-with-interv}.

\subsubsection{Marginal contributions as contrastive statements}
\label{sec:marg-contr-as}

An alternate way to consider Shapley value-based methods as contrastive statements is by examining the marginal contribution of features. The set of \emph{marginal contributions} of each feature $i$, which are averaged in a certain way over all subsets $S$ to calculate the Shapley value, can be thought of as a set of contrastive explanations. Each quantity $\Delta(i,S)$ represents a contrastive explanation for why feature $i$ is important: ``Why choose a model with $S$ and $i$ rather than a model with just $S$? Because it improves $v$ by $\Delta(i,S)$ amount." This quantity is an important part of \emph{stepwise selection}, a modeling procedure in which features which increase the \emph{accuracy} of a model are successively added to the modeling set.

Note that regardless of what order features were actually added to the model in, all permutations are considered when the Shapley value is calculated. It is not clear that taking an average of quantities representing ``all possible contrastive explanations" for a certain set of foils is a sensible way to summarize information. Instead, \citet{miller2019explanation} argues that humans are \emph{selective} about explanations: certain contrasts are more meaningful than others. An example of this is the difference between \emph{necessary} and \emph{sufficient} causes:

\begin{displayquote}
Lipton argues that necessary causes are preferred to sufficient causes. For example, consider mutations in the DNA of a particular species of beetle that cause its wings to grow longer than normal when kept in certain temperatures. Now, consider that there are two such mutations, $M_1$ and $M_2$, and either is sufficient to cause the mutation. To contrast with a beetle whose wings would not change, the explanation of temperature is preferred to either of the mutations $M_1$ or $M_2$, because neither $M_1$ nor $M_2$ are individually necessary for the observed event; merely that either $M_1$ or $M_2$. In contrast, the temperature is necessary, and is preferred, even if we know that the cause was $M_1$.
\end{displayquote}












Consider, without specifying how to quantify the importance $v$ of a feature coalition, computing some kind of allocation for each feature to analyze the positive classification of a beetle with longer wings. Lipton's argument above suggests that since all ``yes" cases share a property $T$, a contrastive statement highlighting this is more relevant than comparisons based on $M_1$ or $M_2$. This is fundamentally at odds with the idea that the ``yes" prediction should be split additively between different coalitions of $M_1$, $M_2$ and $T$, a property induced by the notion of the Shapley value.

\subsection{Using Shapley-valued based methods to enable action}

One motivation for ``explaining" a function is to enable individuals to figure out how to achieve a desirable outcome. For example, one might allow an individual to query the model for a specific contrastive explanation in which the person $p$'s outcome, $f(p)$, is compared with a person $q$ with desirable outcome $f(q) = Q$ determined by the user, such that the user might be able to alter their own situation to approximate $q$. This setup has been formalized as the ``counterfactual explanation" problem by \citet{wachter2017counterfactual} (with an analysis of hidden assumptions by \citet{10.1145/3351095.3372830}). \citet{ustun2019actionable} further specify a way to model this problem by searching for changes within characteristics which are actually mutable; they call this the ``actionable recourse" problem (with a corresponding analysis by \citet{10.1145/3351095.3372876}).

Unlike these methods, Shapley value based frameworks do not explicitly attempt to provide guidance how a user might alter one's behavior in a desirable way. Further, observing that a certain feature carries a large influence over the model does not necessarily imply that changing that feature (even significantly) will change the outcome favorably.


Suppose, in a very simple nonlinear example, that a univariate model is defined as $f(x) = 2 - (x-1)^2$, for some $X \sim N(0,1)$. A person for whom $x = 1$ will get $f(1) = 2$, and $E(X) = 0$, so the Shapley value for this person's single input is then $\phi(x)=2$. Suppose they were hoping for an even higher score. The fact that the value is positive, along with the general knowledge that $1$ is a bit high with respect to an average value of $X$, might make this person think that increasing their $x$ value even more will increase their score -- but it will not.

This problem stems from the fact that the contrastive quantity $E(f(x))$ is not desirable, but even if $v(\emptyset)$ is chosen to be some desirable outcome $f(q)$ of some $q$, such as in \citet{merrick2019explanation}, the Shapley values themselves do not correspond to specific actions: the interventional effect of changing one input from $x$ to that from $q$ is just one of the marginal contributions that are averaged together to form the Shapley value of that input, as we discussed in Section \ref{sec:marg-contr-as}. 

\subsection{Shapley-based explanations for normative evaluation}

Shapley-value-based explanations are primarily used for purposes of normative evaluation: deciding whether a model's behavior is acceptable \cite{bhatt_explainable_2019}. This is done either at the development stage, to help a human evaluate a model, or at the decision-making stage, to help a human evaluate a specific decision made by a model. In this section we explore how the information content of the Shapley value is insufficient for evaluation. We marshal evidence to make three points. Firstly, data scientists do not have a clear mental model of what insights Shapley-value-based analysis brings. Secondly, in the face of this uncertainty, they tend to rely on narrative and confirmation biases. Thirdly, even if they do understand the analysis, it is not obvious that it can be operationalized for specific evaluation tasks. 

Since there is no standard procedure for converting Shapley values into a statement about a model's behavior, developers rely on their own  mental model of what the values represent. \citet{kaur2019interpreting} conducted a contextual inquiry and survey of data scientists to observe their interpretation of interpretability tools including the \texttt{SHAP} Python package. They found that many participants did not have an accurate mental model of what a SHAP analysis represents, yet used them to make decisions on whether the model was ready for deployment, over-trusting and misusing the tool.

Using feature importance during model development in this way is ripe for narrative and confirmation biases. \citet{passi_trust_2018} conducted ethnographic fieldwork with a corporate data science team and described situations in which applying intuition to feature importance was a key component of the model development cycle. In one instance, when developers communicated the results of a modeling effort to project managers, the stakeholders immediately decided it was ``useful'' based entirely on the feature importance list:

\begin{quote}
Certain highly-weighted features matched  business intuitions, and everyone in the  meeting considered this a good thing. 
\dots
Regarding counter-intuitive feature importances, [a data scientist] reminded [the stakeholders] that machine-learning models do not approach data in the same way humans do. He pointed out that models use ``a lot of complex math'' to tell us things that we may not know or fully understand.
\end{quote}


This suggests that even when an individual lacks a correct mental model of the meaning of Shapley values, they may use them to justify their evaluation anyway, whether or not this analysis is well-founded.

In support of this hypothesis, empirical studies have shown that interpretability is not always helpful in task-specific settings.
~\citet{poursabzi2018manipulating}, for instance, demonstrated that ``interpretable" models may not be easier to evaluate:

\begin{quote}Participants who were shown a clear model with a small number of features were better able to simulate the model’s predictions. However, contrary to what one might expect when manipulating interpretability, we found no improvements in the degree to which participants followed the model’s predictions when it was beneficial to do so. Even more surprisingly, increased transparency hampered people’s ability to detect when the model makes a sizable mistake and correct for it, seemingly due to information overload.\end{quote}

This suggests that common intuition for the benefits of interpretability (and the types of questions it can help answer) may be based on faulty assumptions, and these questions should instead be concretely specified and tested. For instance, data scientists might want to know: 

\begin{itemize}
\item Whether an error was made at any point in the data processing pipeline for a certain feature
\item Whether the model is acting upon spurious correlations or other artifacts of training data
\item Whether the model exhibits inappropriate biases
\item Whether the model's accuracy will improve if a certain feature is included or excluded
\end{itemize}

While Shapley-value-based methods might help qualitatively inform investigations that \emph{lead} to answers to these questions, it is not clear that they provide direct answers to any \emph{specific} question related to the points of interest above. ~\citet{weerts2019human}, for instance, conducted a human-grounded evaluation of SHAP and did not find evidence that it helped users assess the correctness of predictions.



\section{Conclusion}
\label{sec:concl}

Shapley values enjoy mathematically satisfying theoretical properties as a solution to game theory problems. However, applying a game theoretic framework does not automatically solve the problem of feature importance, and our work shows that in fact this framework is ill-suited as a general solution to the problem of quantifying feature importance. Rather than relying on notions of mathematical correctness, our work suggests that we need more focused approaches that stem from specific use cases and models, developed with human accessibility in mind. 

\paragraph{Acknowledgments.}
\label{sec:acknowledgements}

This research was supported in part by the National Science Foundation under grants IIS-1633724, IIS-1633387, DMR-1709351, IIS-1815238, the DARPA SD2 Program, and the ARCS Foundation.


\bibliography{example_paper}
\bibliographystyle{icml2020}


\end{document}